\documentclass[11pt]{article}

\usepackage{microtype} 
\usepackage{booktabs}  
\usepackage{url}  

\usepackage{amsmath}
\usepackage{amsthm}

\usepackage[english]{babel}
\usepackage{blindtext}

\usepackage[final]{mais}

\usepackage[square,numbers]{natbib}
\usepackage{xspace}
\usepackage{tabularx}

\usepackage{multirow}
\usepackage{makecell}
\usepackage{booktabs, colortbl}       
\usepackage{amsfonts}       
\usepackage{nicefrac}       
\usepackage{microtype}      

\usepackage{csquotes}
\usepackage{latexsym}
\usepackage{graphicx}
\usepackage{placeins}
\usepackage{bm}
\usepackage{balance}

\usepackage{subcaption}
\usepackage{caption}

\usepackage{mathtools}
\usepackage{hyperref}

\usepackage{pifont}
\usepackage{wrapfig}

\newcommand\cl{\texttt{CL}\xspace}
\newcommand\camcos{\texttt{CAM-COS}\xspace}  
\newcommand\attcos{\texttt{ATT-COS}\xspace}  
\newcommand\aus{\texttt{AUs}\xspace}
\newcommand\au{\texttt{AU}\xspace}
\newcommand\rafdb{\texttt{RAF-DB}\xspace}
\newcommand\affectnet{\texttt{AffectNet}\xspace}

\definecolor{Gray}{gray}{0.85}
\definecolor{darkergreen}{RGB}{21, 152, 56}

\newcolumntype{g}{>{\columncolor{Gray}}c}
\newcommand{\reals}{\mathbb{R}}

\newcommand{\normx}[1]{\ensuremath \lVert#1\rVert}
\newcommand{\edit}[1]{\textcolor{black}{#1}}

\title{Spatial Action Unit Cues for Interpretable Deep Facial Expression Recognition}

\author[1]{\nameemail{Soufiane Belharbi}{soufiane.belharbi@etsmtl.ca}}
\author[1]{\nameemail{Marco Pedersoli}{Marco.Pedersoli@etsmtl.ca}}
\author[1]{\nameemail{Alessandro Lameiras Koerich}{Alessandro.LameirasKoerich@etsmtl.ca}}
\author[2,3]{\nameemail{Simon Bacon}{Simon.Bacon@concordia.ca}}
\author[1]{\nameemail{Eric Granger}{Eric.Granger@etsmtl.ca}}

\affil[1]{LIVIA, Dept. of Systems Engineering, ÉTS, Montreal, Canada}
\affil[2]{Dept. of Health, Kinesiology \& Applied Physiology, Concordia Univ., Montreal, Canada}
\affil[3]{Montreal Behavioural Medicine Centre, Montreal, Canada \newline
\texttt{\{soufiane.belharbi,eric.granger\}@etsmtl.ca}
}

\hypersetup{%
  pdfauthor={Belharbi et al.}, 
  pdftitle={Spatial Action Unit Cues for Interpretable Deep Facial Expression Recognition},
  pdfsubject={Facial expression recognition and interpretability},
  pdfkeywords={AI Symposium 2024, October 18th 2024, Montréal}
}

\begin{document}

\maketitle

\begin{figure}[htp!]
\centering
  \centering
  \includegraphics[width=0.36\linewidth]{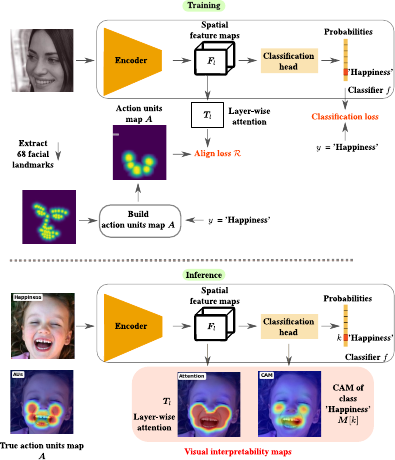}
  \includegraphics[width=0.47\linewidth]{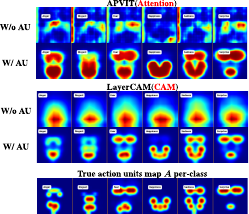}
  \caption[Caption]{
  \footnotesize
  \textbf{Left}: Our interpretable classifier for the FER task (training and inference). 
  \textbf{Right}: 
  Illustration of per-class average {\color{red}attention} and {\color{red}CAM} maps over all test set of \rafdb with and without \au alignment. Expressions from left to right: 'Anger', 'Disgust', 'Fear', 'Happiness', 'Sadness', 'Surprise'.}
  \label{fig:rafdb-att-cam}
\end{figure}

\noindent\textbf{Introduction:} Facial expression recognition (FER) has recently received much interest in the computer vision and machine learning communities~\citep{ZhengM023}. 
FER remains challenging, particularly for real-world applications. This is due to the subtle differences between expressions, leading to low inter-class variability. To tackle these challenges, many deep learning methods have been proposed to train highly accurate classification models~\citep{BonnardDDB22,xue2022vision, ZhengM023}.
However, end-users may not require only an accurate classifier that yields a classification score in multiple critical applications~\citep{deramgozin22}. They may also need the model to provide an interpretable decision~\citep{yadav22}. While existing works excel at classification performance, they lack interpretability. 
This work aims to provide a learning strategy to train deep model to perform classification while being interpretable. Experts commonly rely on a codebook of AUs to determine a basic facial expression~\citep{Martinez19} which make them a reliable interpretation by the model. Therefore, we propose to leveraging spatial action units as guidance to train interpretable classifiers (Fig.\ref{fig:rafdb-att-cam}).  In addition, we exploit Class Activation Mapping methods (CAMs)~\citep{murtaza24,rony2023deep} for the first time in FER models for further interpretability.

\begin{figure*}[htp!]
\centering
  \centering
  \includegraphics[width=0.8\linewidth]{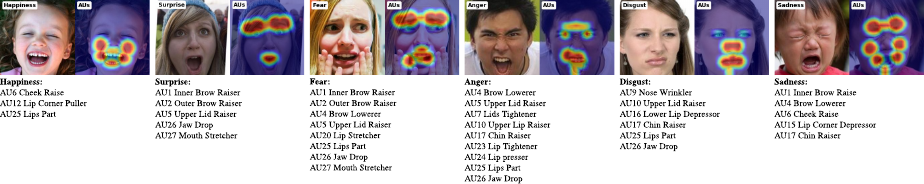}
  \caption[Caption]{
  \footnotesize
  Codebook of basic facial expressions and their associated \aus~\citep{Martinez19}.
  }
  \label{fig:codebook-fer-aus}
\end{figure*}

\noindent\textbf{Proposed Method:} Our proposed method is presented in Fig.\ref{fig:rafdb-att-cam}. During training, we first build a discriminative spatial \aus map ${\bm{A} \in \reals^{h\times w}}$ per image using its image-class label, and the codebook in Fig.\ref{fig:codebook-fer-aus}. The spatial \au map is built using the image expression to select the right corresponding \au subset in combination with facial landmarks, which are employed to localize these same \aus in the image. In particular, the location of landmarks is used to estimate \au positions. For instance, the right 'Cheek' location is estimated using landmark 47 (middle of the low right eye) and 11 (right side of the jaw). This discriminative map ${\bm{A}}$ will be used only during training.

To incorporate this localization prior into a deep model, we use its layer-wise attention ${\bm{T}_l}$ at layer $l$, and aim to align it with the \au map ${\bm{A}}$ using cosine similarity as follows:
\begin{equation}
    \bm{T}_l = \frac{1}{n}\sum_{j = 0}^n \bm{F}_l[j] , \quad \mathcal{R}(\bm{T}_l, \bm{A}) = \frac{\sum \left(\bm{T}_l \odot \bm{A} \right) }{\normx{\bm{T}_l}_2 \normx{\bm{A}}_2}\;.
    \vspace{-4mm}
\end{equation}
The full model is trained to perform classification, and localization simultaneously as follows:
\begin{equation}
    \label{eq:total-loss}
    \min_{\bm{\theta}} \quad - \log(f(\bm{X};\bm{\theta})_y) + \lambda (1 - \mathcal{R}(\bm{T}_l, \bm{A}))\;.
    \vspace{-6mm}
\end{equation}

The computational cost added by the alignment term is negligible since it can be easily computed on GPU. This leads to a training time similar to the vanilla case, where no alignment is used. The only required pre-processing is facial landmark extraction, which can be done offline once and stored on disk. \au map ${\bm{A}}$ can be computed in a negligible time on CPU on the fly during training using the image-class label only as supervision.
Since image-class labels are unavailable at test time, \au maps can not be built using the lookup table emotion-\aus. Hence, training the model to produce them along with the true label as we propose is more realistic and practical. 

\noindent\textbf{Experimental Results:} We evaluate our method on 2 public FER benchmarks: \rafdb~\citep{li17}, and \affectnet~\citep{Li21}. Tab.\ref{tab:cl-cam-attloc} shows classification and localization performance. Visual results of localization are illustrated in Fig.\ref{fig:rafdb-att-cam}. These results show that using our proposal improves both classification and localization (interpretability). 
The full version of this work can be found at~\citep{belharbi24-fer-aus}. Our code is publicly available at {\href{https://github.com/sbelharbi/interpretable-fer-aus}{\color{red}{github.com/sbelharbi/interpretable-fer-aus}}}.

{
\setlength{\tabcolsep}{3pt}
\renewcommand{\arraystretch}{1.1}
\begin{table*}[ht!]
\centering
\resizebox{.73\textwidth}{!}{%
\centering
\small
\begin{tabular}{|l|ccc|cc|ccc|cc|cc|cc|}
\cline{1-15}
&& \multicolumn{6}{c}{\rafdb} && \multicolumn{6}{c|}{\affectnet} \\
\cline{2-15}
&& \multicolumn{2}{c}{\cl} & \multicolumn{2}{c}{\camcos} & \multicolumn{2}{c}{\attcos} && \multicolumn{2}{c}{\cl} & \multicolumn{2}{c}{\camcos} & \multicolumn{2}{c|}{\attcos} \\
\cline{2-15}
\textbf{Method}  && w/o \au & w/ \au &  w/o \au & w/ \au &  w/o \au & w/ \au &&  w/o \au & w/ \au &  w/o \au & w/ \au &  w/o \au & w/ \au \\
\cline{1-15} 
\textbf{CNN-based}  &&  \multicolumn{13}{c|}{} \\
CAM~\citep{zhou2016learning} {\small \emph{(cvpr,2016)}}                     && $88.20$  & $\bm{88.95}$  & $0.55$  & $\bm{0.70}$ & $0.57$  & $\bm{0.85}$  && $60.88$ & $\bm{62.37}$ & $0.56$  & $\bm{0.69}$ & $0.64$  & $\bm{0.82}$\\
WILDCAT~\citep{durand2017wildcat} {\small \emph{(cvpr,2017)}}                && $88.26$  & $\bm{88.85}$  & $0.52$  & $\bm{0.69}$ & $0.47$  & $\bm{0.85}$  && $59.88$ & $\bm{61.62}$ & $0.62$  & $\bm{0.80}$ & $0.61$  & $\bm{0.81}$\\
GradCAM~\citep{SelvarajuCDVPB17iccvgradcam} {\small \emph{(iccv,2017)}}      && $88.39$  & $\bm{88.85}$  & $0.55$  & $\bm{0.74}$ & $0.63$  & $\bm{0.85}$  && $60.77$ & $\bm{62.08}$ & $0.53$  & $\bm{0.75}$ & $0.65$  & $\bm{0.82}$\\
GradCAM++~\citep{ChattopadhyaySH18wacvgradcampp} {\small \emph{(wacv,2018)}} && $87.84$  & $\bm{89.14}$  & $0.60$  & $\bm{0.82}$ & $0.52$  & $\bm{0.87}$  && $60.22$ & $\bm{62.45}$ & $0.66$  & $\bm{0.83}$ & $0.65$  & $\bm{0.82}$\\
ACoL~\citep{ZhangWF0H18} {\small \emph{(cvpr,2018)}}                         && $87.94$  & $\bm{88.68}$  & $0.54$  & $\bm{0.67}$ & $0.46$  & $\bm{0.84}$  && $58.28$ & $\bm{61.48}$ & $0.55$  & $\bm{0.65}$ & $0.60$  & $\bm{0.81}$\\
PRM~\citep{ZhouZYQJ18PRM}  {\small \emph{(cvpr,2018)}}                       && $88.13$  & $\bm{88.88}$  & $0.48$  & $\bm{0.59}$ & $0.43$  & $\bm{0.85}$  && $57.77$ & $\bm{60.97}$ & $0.52$  & $\bm{0.75}$ & $0.55$  & $\bm{0.82}$\\
ADL~\citep{ChoeS19} {\small \emph{(cvpr,2019)}}                              && $87.45$  & $\bm{88.65}$  & $0.50$  & $\bm{0.63}$ & $0.51$  & $\bm{0.85}$  && $57.88$ & $\bm{61.25}$ & $0.54$  & $\bm{0.66}$ & $0.65$  & $\bm{0.83}$\\
CutMix~\citep{YunHCOYC19} {\small \emph{(eccv,2019)}}                        && $88.39$  & $\bm{88.59}$  & $0.55$  & $\bm{0.57}$ & $0.51$  & $\bm{0.80}$  && $58.74$ & $\bm{59.88}$ & $0.56$  & $\bm{0.58}$ & $0.57$  & $\bm{0.82}$\\
LayerCAM~\citep{JiangZHCW21layercam} {\small \emph{(ieee,2021)}}             && $87.90$  & $\bm{88.88}$  & $0.60$  & $\bm{0.84}$ & $0.52$  & $\bm{0.86}$  && $60.77$ & $\bm{62.45}$ & $0.66$  & $\bm{0.83}$ & $0.65$  & $\bm{0.82}$\\
\cline{1-15} 
\textbf{Transformer-based}  &&  \multicolumn{13}{c|}{} \\
TS-CAM~\citep{gao2021tscam} {\small \emph{(iccv,2021)}}                      && $86.70$  & $\bm{88.00}$  & $0.58$  & $\bm{0.71}$ & $0.55$  & $\bm{0.88}$ && $58.99$ & $\bm{59.54}$ & $0.57$  & $\bm{0.58}$ & $0.48$  & $\bm{0.79}$\\
APViT~\citep{xue2022vision} {\small \emph{(ieee,2022)}}                      && $91.00$  & $\bm{91.03}$  & $--$    & $--$        & $0.38$  & $\bm{0.85}$ && $60.62$ & $\bm{62.28}$ & $--$    & $--$              & $0.45$  & $\bm{0.84}$\\
\cline{1-15} 
\end{tabular}
}
\caption{
\footnotesize
\textbf{Classification (\cl), CAM-localization (\camcos)}, and \textbf{Attention-localization (\attcos)} (at layer 5) performance on \rafdb and \affectnet test sets with and without \aus across methods.}
\label{tab:cl-cam-attloc}
\end{table*}
}

\FloatBarrier

\section*{Acknowledgments}
\edit{This work was supported in part by the Fonds de recherche du Québec – Santé (FRQS), the Natural Sciences and Engineering Research Council of Canada (NSERC), Canada Foundation for Innovation (CFI), and the Digital Research Alliance of Canada.
}


\bibliography{references}



\end{document}